\pdfoutput=1

\documentclass[11pt,a4paper]{article}
\usepackage{acl}

\usepackage{times}
\usepackage{latexsym}
\usepackage{arydshln}
\usepackage{comment}

\usepackage{xcolor}
\usepackage{soul}

\usepackage[T1]{fontenc}

\usepackage[utf8]{inputenc}

\usepackage{microtype}

\usepackage{booktabs}
\usepackage{graphicx}
\usepackage{subcaption}
\usepackage{xspace}
\usepackage{multirow}

\usepackage{enumitem, array}
\newcolumntype{R}[1]{>{\raggedleft\arraybackslash}p{#1}}
\newcolumntype{L}[1]{>{\raggedright\arraybackslash}p{#1}}
\newcolumntype{C}[1]{>{\centering\let\newline\\\arraybackslash\hspace{0pt}}m{#1}}

\newcommand{\ours}{\texttt{\textsc{WeSEEL}}\xspace}
\newcommand{\ccme}{\texttt{\textsc{CCME}}\xspace}
\newcommand{\ccel}{\texttt{\textsc{CCEL}}\xspace}

\title{Weakly Supervised Medical Entity Extraction and Linking \\ for Chief Complaints}

\author{
  Zhimeng Luo, Zhendong Wang, Rui Meng, Diyang Xue, Daqing He \\
  School of Computing and Information, University of Pittsburgh, Pittsburgh, PA, United States 
  \AND
  Adam Frisch \\
  Department of Emergency Medicine, University of Pittsburgh, Pittsburgh, PA, United States \\
}

\begin{document}




\maketitle
\begin{abstract}
A Chief complaint (CC) is the reason for the medical visit as stated in the patient's own words. It helps medical professionals to quickly understand a patient's situation, and also serves as a short summary for medical text mining. However, chief complaint records often take a variety of entering methods, resulting in a wide variation of medical notations, which makes it difficult to standardize across different medical institutions for record keeping or text mining. 
In this study, we propose a weakly supervised method to automatically extract and link entities in chief complaints in the absence of human annotation. We first adopt a split-and-match algorithm to produce weak annotations, including entity mention spans and class labels, on 1.2 million real-world de-identified and IRB approved chief complaint records.
Then we train a BERT-based model with generated weak labels to locate entity mentions in chief complaint text and link them to a pre-defined ontology.
We conducted extensive experiments and the results showed that our Weakly Supervised Entity Extraction and Linking (\ours) method produced superior performance over previous methods without any human annotation.

\end{abstract}






\section{Introduction}
A chief complaint (CC) is an initial statement of patient derived medical issues, which is often elicited prior to formal medical tests and diagnoses. It provides a brief statement about a patient's reasons for encounter, current symptoms, and medical history~\cite{chang2020generating}. It can be of great help to medical professionals in understanding a patient's situation and lead to the appropriate diagnoses and treatments~\cite{wagner2006chief}.
Furthermore, it can be seen as a summary of patient profiles and medical records, and it is useful for a wide variety of medical text mining tasks. But the application of chief complaints is greatly restricted by the fact that there lacks a widely accepted standard for data entry of chief complaints. Hospitals and health care systems adopt different ontology and standards to enter and store chief complaint data~\cite{horng2019consensus}, which causes chief complaint records contain various local terminologies and abbreviations.

Commonly, a chief complaint record is a piece of free-text statement and may contain one or multiple medical entity mentions. In figure~\ref{fig:cc_examples}, several chief complaint records and the corresponding concept annotations in HaPPy ontology~\cite{horng2019consensus} are shown, which also stress the typical difficulties in understanding chief complaints. 
First of all, synonyms are very common in chief complaints. The same concept can appear in different forms, e.g. ``chest pain'', ``CP'', or ``cerebral palsy''. Another common issue is word sharing. For example, in the record of ``migraine with neck/back pain, fever'', the word ``pain'' is shared by both ``neck pain'' and ``back pain''. Also, in a more complex example ``chills/body aches/n/d/r side jaw pain'', the slash ``/'' acts not only as the separator to split multiple entities, but also a part of the abbreviation ``n/d'' (i.e. nausea and diarrhea). In these cases, it is even challenging for domain experts to divide and identify entities. Moreover, a chief complaint record is often entered as a piece of free text, where misspellings can occur commonly. All these examples exhibit the challenges in processing chief complaints with automatic NLP techniques.

\begin{figure*}
\fontsize{9}{12}\selectfont
\centering
\begin{tabular}{lll}
\toprule
\textbf{Chief complaint Records} & \textbf{Concepts in Ontology} & \textbf{NLP Challenges} \\ 
\midrule
urinary tract pain & dysuria & Synonym understanding \\
migraine with neck/back pain, fever & migraine, neck pain, back pain, fever & Resolving shared tokens \\
chills/body aches/n/d/r side jaw pain & chills, body aches, nausea, diarrhea, jaw pain & Segmentation \& abbreviations \\
ha light headed fatigue r arm pain & headache, dizziness, fatigue, arm pain & Missing separators \\
\bottomrule
\end{tabular}
\vspace{-0.5em}
\caption{Examples of chief complaint records and corresponding concepts in HaPPy ontology.}
\vspace{-1em}
\label{fig:cc_examples}
\end{figure*}

Due to the difference in terms of data format (long vs. short), target applications (general purpose vs. emergency) and concept ontology (the number of concepts general medical records is much larger), many tools designed for general medical domain cannot be directly applied to CC documents.
Several attempts have been made to automate the process of mining entities in chief complaints. For example, \citeauthor{Karagounis2020Coding} utilized rule-based matching tool MetaMap \cite{aronson2001effective} to map free-text chief complaints to ICD-10-CM codes. 
\citeauthor{chang2020generating} fine-tuned a BERT model on 1.8 million emergency department (ED) chief complaint records to derive a domain-specific representation, which achieved improved performance on predicting chief complaint concepts. Despite these advancements, most existing efforts \cite{DARA2008613,conway2013using,duangsuwan2018semi,lee2019chief,valmianski2019evaluating,hsu2020natural,Osborne2020Identification} on automatic chief complaint processing viewed the task as a classification problem, whereas in the real-world case, one chief complaint record often contains multiple entities (e.g. the examples shown in Figure~\ref{fig:cc_examples}). Although multiple entities can be formulated as a multi-label classification as well, locating the span of entities and linking them to ontology is a more direct and explainable way than multi-label classification.
 Consequently, 
it is more appropriate and useful to view the task as entity extraction problem (i.e., identifying the actual mention span of each entity in a free text) and entity linking problem (i.e., linking each entity mention to an entry in a chief complaint ontology).

Besides, nowadays most machine learning applications require extensive annotated data to train their models~\cite{hsu2016data,liao2015development,rochefort2015037,shen2017leveraging}. 
However, in the case of medical domain, the data annotation is prohibitively sensitive and expensive, due to the data privacy concern and high cost of recruiting healthcare professionals. The lack of high-quality annotation also hinders the progress of NLP methods and applications for chief complains. This becomes more serious in the domain of health care due to the facts that publicly available clinical corpora is rare, and the requirement of clinical knowledge to be able to correctly annotated clinical text. Even worse, the annotation of CC requires the annotators have to be nurse or doctor who works in the front line of ED, in order to familiar with the common abbreviations and misspelling in the health care system. Therefore, state-of-the-art machine learning models that require large amount of labeled training data are not applicable for CC corpus.

In this study, we aim to advance the application of NLP to chief complaints by proposing a novel task setup consisting of two steps: extracting entity mentions from a chief complaint record and linking them to a given ontology. We utilize a BERT-based extraction and linking model to address this task and propose a split-and-match (S\&M) algorithm to generate weak annotations to resolve the shortage of annotated data. 
We conducted experiments with 1.2 million free-text ED chief complaint records from University of Pittsburgh Medical Center (UPMC) and the results show that the proposed method can achieve better performance compared with various baselines.

Specifically, the contributions of our work are:

\begin{itemize}
\item 
To our knowledge, this is the first study mining entities in chief complaints with two explicit steps (extraction and linking), which is more direct and explainable than the classification setup in previous studies.
\item We propose a weak supervision method \ours for extracting and linking entities in chief complaints. We demonstrate the superiority of our method over various baselines with extensive experiments and analyses.
\item We contribute a new dataset, containing 1.2 million free-text chief complaint records from emergency departments of local hospitals, and 1,013 data examples are manually annotated by clinicians for the purpose of testing. 
\end{itemize}


\begin{figure*}
\begin{subfigure}[b]{1.0\textwidth}
  \centering
  \includegraphics[width=0.9\textwidth]{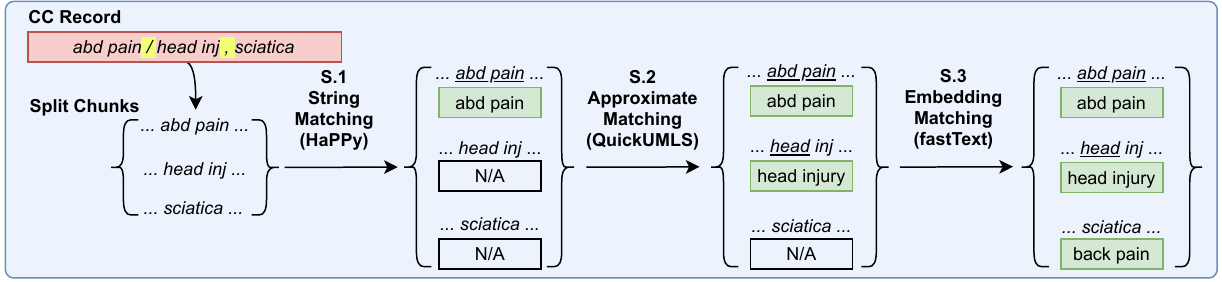}
  \caption{Process flow of weak label generation. Three examples are shown and successfully matched to concepts in the ontology at different stages (indicated in green box).}
  \label{fig:weak_label_pipeline}
\end{subfigure}

\begin{subfigure}[b]{1.0\textwidth}
  \centering
  \includegraphics[width=0.9\linewidth]{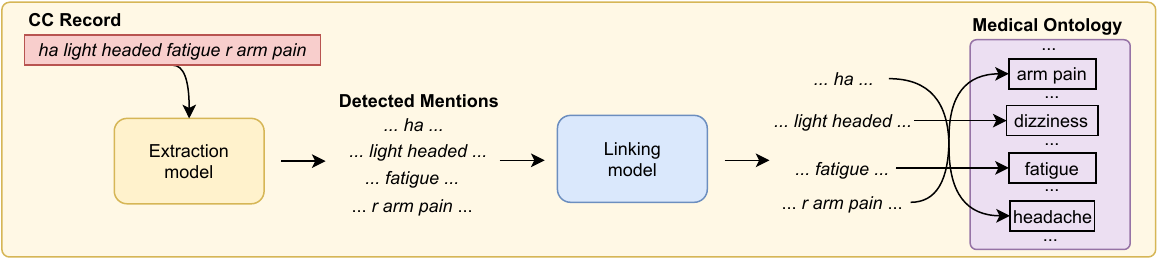}
  \caption{Schematic diagram of our model for entity extraction and linking in chief complaint.}
  \label{fig:model_pipeline}
\end{subfigure}
\vspace{-2em}
\caption{Overview of our proposed method \ours.}
\vspace{-1em}
\label{fig:framework}
\end{figure*}

\section{Related Works}

\subsection{Studies on Chief Complaint}


Previous works on automatic CC processing tried to link entities from CC records to a given ontology.
Single-label classification was performed on 2.1 million patient-level ED visit records with ICD codes \cite{lee2019chief}. \citet{valmianski2019evaluating} compared embedding from BERT and its variants BioBERT \cite{lee2020biobert} and ClinicalBERT \cite{huang2019clinicalbert} with TF-IDF on a dataset of ~200,000 patient-generated reasons-for-visit entries and mapped them to 795 unique chief complaint concepts. \citet{teng2020automatic} formulated automatic diagnose code assignment as multi-label classification to predict ICD codes from free-text medical records including chief complaints. Most datasets in these studies are not publicly available.


\subsection{Medical Entity Extraction and Linking}
Named Entity Recognition (NER) is a common NLP technique for locating entity mentions in a text.
\citet{zhao2019neural} developed a multi-task framework to jointly tackle the task of medical NER and entity normalization. \citet{bhatia2019comprehend} developed a web service for medical NER and relationship extraction in medical data. To tackle the issue of limited data annotation, \citet{hofer2018few} proposed a method for few-shot learning of NER in medical domain.

Entity linking (EL) aims to map entity mentions in text to concepts in a knowledge base or ontology. String matching- and rule-based systems like cTAKES~\cite{savova2010mayo} and MetaMap~\cite{aronson2001effective} were proposed to link medical entities in EHRs. Recent studies also utilized neural networks. 
MedType~\cite{vashishth2020medtype} was pretrained with large auto-annotated datasets and achieved state-of-the-art performance on multiple medical entity linking benchmarks. MedCAT \cite{Kraljevic2021-ln} was proposed to extract information from EHR and link to biomedical ontologies.
~\citet{ji2020bert} proposed to treat the task of entity normalization as ranking.
~\citet{Chen_Varoquaux_Suchanek_2021} applied BERT to learn alignment between mentions and entity names. However, most of those studies require a fixed list of mention candidates, which adding an extra retrieval step, make the framework more complicated and may further introduce noises.

\subsection{Resolving Shortage of Annotated Data}

Weakly supervised method is effective in training machine learning models with automatically generated pseudo-labeled data. It has been utilized in sentiment analysis~\cite{severyn2015twitter}, relation extraction~\cite{hoffmann2011knowledge,bing2015improving}, and information retrieval~\cite{dehghani2017neural}. It is also used in medical topics such as drug-drug interactions~\cite{li2016topic}, medical term identification~\cite{neveol2017making}, and sentence extraction in clinical trial reports~\cite{wallace2016extracting}.


\section{Methods}
Formally, chief complaint entity extraction and linking can formulate as follows. A set of medical concepts in an ontology is denoted as $C = \{c_1, c_2, ..., c_N\}$ and a free-text chief complaint record is a sequence of words $D = (w_1, w_2, ..., w_{|D|})$. The goal of the task is to (1) identify one or multiple entity mentions $M = \{m_1, m_2, ..., m_{|M|}\}$, each indicated by an index span $(w_{i},...,w_j)$ of $D$, and (2) link them to corresponding concepts $c \in C$ in the ontology.


To reduce the reliance of human annotation, we utilize a split-and-match algorithm to generate weak training labels 
for both extraction and linking. 
We build a sequence labeling model with BERT (\ccme-BERT) as the extraction model to identify chief complaint entity mentions.
Then 
a BiLSTM model (\ccel) is adopted as the entity linking model to combine the word-level and character-level embedding of entity mentions, followed by a feedforward neural network to predict concept label in an ontology. 
The schematic diagram of the proposed label generation process and the entity recognition model is shown in Figure~\ref{fig:framework}.

\setcounter{subsection}{-1}
\subsection{Pre-processing Chief Complaint Records}
\label{sec:preprocess}
We observe that a large percent of chief complaint records (55.7\% on labeled subset) contain multiple symptoms or reasons, and triage nurses often use punctuation marks to delimit multiple parts. Thus we manually select 10 common punctuation marks that are used as separators.




\subsection{Generating Weak Labels for Training}
\label{sec:weak_label_gen}
Weak supervision trains machine learning models with noisy sources to provide supervision signals. 
In our case, a large proportion of chief complaint records contain explicit punctuation separators that can segment a chief complaint record into multiple chunks and some of them can be chief complaint entity mentions. Thus, we propose a split-and-match algorithm to automatically generate weak labels of chief complaint entity mentions and concept labels. 

\begin{enumerate}
    \item \textbf{Split}: We split a chief complaint record into multiple chunks by pre-defined separators.
    \item \textbf{Match}: For each text chunk, we 
    check if 
    it can match to a concept in the ontology by various methods. Matched chunks will be saved as weak annotations for training our models.
\end{enumerate}

Matching chief complaint mentions to corresponding concepts in an ontology is the core of this algorithm. To seek a balance between precision and recall of the weak labels, we employ three matching methods as a pipeline, as illustrated in Figure~\ref{fig:weak_label_pipeline}: 
\textit{S.1 - Exact string match}. We simply check if a chunk exactly matches to any alternative form of a concept in HaPPy ontology; 
\textit{S.2 - Approximate string matching}. We use QuickUMLS~\cite{soldaini2016quickumls}, a tool for 
extracting medical concepts using an approximate dictionary matching algorithm. This helps to resolve misspelling and lexical variations. 
\textit{S.3 - Embedding-based matching}. It is achieved by computing the cosine similarity between the embedding of the chunk and that of an ontology concept. The embedding is obtained with fastText~\cite{joulin2016fasttext}, trained on our chief complaint corpus. 
In this way, a large amount of ``weak'' annotations of entity spans and concept labels are collected. 

\subsection{Entity Mention Extraction}

We formulate the task of detecting entity mentions from a chief complaint record as a sequence labeling problem following the BIO tagging scheme, which is common 
in NER tasks. Each token takes a label, where
``B/I'' denotes a beginning/inside token of entity mentions, and ``O'' denotes other tokens outside mentions or separator tokens.
We refer to our models for Mention Extraction in chief complaints as \ccme.


One important feature of our model is the usage of the label smoothing technique to counter the noise in generated weak labels. With the help of label smoothing~\cite{szegedy2016rethinking}, the model can be aware of the qualify of target labels so it can learn accordingly. 
We adjust the label smoothing to accommodate the weak span labels since the similarity score in matching indicates the confidence level of a weak label. 


\subsection{Linking Entities to Ontology}

\begin{figure}[!htbp]
    \centering
    \includegraphics[width=0.45\textwidth]{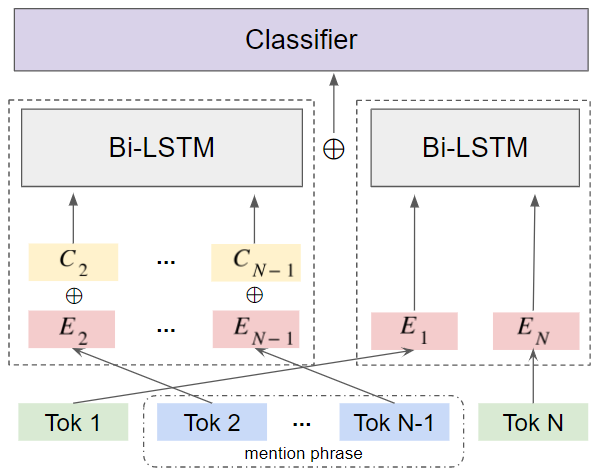}
    \caption{Architecture of the model for entity linking in chief complaints (\ccel) .}
    \label{fig:model2}
\end{figure}

Each extracted entity mention will be linked with a given ontology through a classification model. We concatenate both word embedding and character embedding of an entity mention as the input, to accommodate the variable forms of medical entities. BERT embedding is not considered here as it's hard to be integrated with character embedding.
Besides, two BiLSTMs are used to encode the tokens in its context from both directions, since the context before/after the mention may capture different information.
Lastly, the outputs from both BiLSTMs are concatenated and fed to a softmax layer to obtain the concept label in the ontology. The diagram of our entity linking model called \ccel is shown in Figure~\ref{fig:model2}. 

\begin{table*}[h]
\fontsize{9}{11}\selectfont
\centering
\begin{tabular}{lcccccc}
\toprule
& \multicolumn{3}{c}{\textbf{Partial Match}} & \multicolumn{3}{c}{\textbf{Exact Match}}\\
\textbf{Models} & \textbf{Precision} & \textbf{Recall} & \textbf{F1} & \textbf{Precision} & \textbf{Recall} & \textbf{F1} \\ 
\midrule
\dag S\&M (S.1) & \underline{95.81} & 36.90 & 53.28 & \underline{92.69} & 35.70 & 51.54 \\
\dag S\&M (S.1 + S.2) & 81.15 & 46.04 & 58.75 & 67.46 & 38.28 & 48.84 \\
\dag S\&M (S.1 + S.2 + S.3) & 69.64 & \textbf{57.36} & 62.91 & 55.66 & 45.84 & 50.27 \\
\midrule
\ccme-LSTM & 78.45 & 48.26 & 59.76 & 67.25 & 42.39 & 52.00 \\
\ccme-BERT & 81.37 & 53.43 & 64.50 & 71.46 & 48.52 & 57.80 \\
\ccme-BERT (soft) & 83.41 & \underline{56.70} & \textbf{67.51} & 72.95 & \textbf{49.59} & \underline{59.04} \\
\ccme-ClinicalBERT (soft) & 83.35 & 56.46 & \underline{67.32} & 72.94 & \underline{49.41} & 58.91 \\
\ccme-BioBERT (soft) & 81.64 & 54.25 & 65.18 & 71.25 & 48.38 & 57.63 \\
\ccme-PubMedBERT (soft) & 84.14 & 54.61 & 66.23 & 73.27 & 48.73 & 58.53 \\
\midrule
\ccme-BERT (soft) + S\&M (S.1) & \textbf{96.28} & 44.86 & 61.20 & \textbf{92.94} & 43.30 & \textbf{59.08} \\
\ccme-BERT (soft) + S\&M (S.1 + S.2) & 86.13 & 51.82 & 64.71 & 76.86 & 46.24 & 57.74 \\
\bottomrule
\end{tabular}
\caption{Entity extraction performance of different models. Scores are computed in \textbf{Partial Match} and \textbf{Exact Match} mode of SemEval'13. The \textbf{best}/\underline{2nd-best} scores in each column are in bold/underlined. \dag S.1, S.2, S.3 refer to string matching algorithms in Figure \ref{fig:framework}(a), respectively.}
\vspace{-1em}
\label{table:entity_extraction}
\end{table*}

\section{Experimental Setup}
\subsection{Data}
Our dataset\footnote{All code/data for reproducing the results will be released at \url{https://github.com/anonymous_repo}, under IRB restrictions.} contains 1,232,899 free-text chief complaint records (in English) of patients' visits at 15 emergency departments of local healthcare institutes, including critical access hospitals, community hospitals, and tertiary care referral centers. All these departments use the same electronic health record system, but do not mandate a specific data entry format. The time span of the data is from 2015 to 2017. The dataset has been de-identified and approved by IRB. 

HaPPy ontology is utilized as the target ontology to link entities. Additionally, we reduce the size of HaPPy ontology to 501 medical concepts (originally 692) by removing child nodes that have no significant clinical difference with their parent node. 
For example, two child nodes ``ruq abdominal pain'' (``ruq'' means right upper quadrant) and ``rlq abdominal pain'' (``rlq'' means right lower quadrant) are merged to their parent node ``abd pain''. 

The testset consists of 1,013 instances (randomly sampled from the original collection) and 1,771 chief complaint mentions with 211 unique concepts. Two domain experts independently annotated the data (locating the chunk of mentions and linking mentions to concepts in HaPPy ontology), whereas the 2nd expert only annotated 200 data points to measure the inter-annotator agreement. There were no weak labels used in the process of human annotation. The resulting Cohen's Kappa of concept classification (entity type) is 0.9326, and the accuracy of exact span overlap is 0.9029, both indicating a very high reliability between two annotators.


\subsection{Experiment Settings}
With regard to the entity mention extraction, we use weak labels generated from the split-and-match (S\&M) algorithm introduced in Sec~\ref{sec:weak_label_gen}: S\&M (HaPPy), S\&M (QuickUMLS), and S\&M (Embedding) as three weak baselines.
Besides, we also experiment with LSTM and several variants of BERT for \ccme: \ccme-LSTM, \ccme-BERT, \ccme-ClinicalBERT, \ccme-BioBERT and \ccme-PubMedBERT \cite{gu2021domain}. We refer to the \ccme-BERT model with the label smoothing as \ccme-BERT (soft).

As for entity linking, a fastText- and a BERT-based classification model are used as baseline models. The input to fastText and BERT is the same as \ccel, i.e. mention tokens with context tokens within a window size of 2 (two tokens before/after the mention). 
To compare the performance of adding entity titles as input, BERT-based Ranking \cite{ji2020bert} has experimented as entity linking model. To compare task settings, We train a fastText model following the single-label paradigm, that only predicts one concept class for each chief complaint record. We take MedType~\cite{vashishth2020medtype} (use QuickUMLS for extraction) as a baseline system.

We follow the setting of \mbox{\cite{yang2018design}} for training the \ccme-LSTM model and default settings of HuggingFace \mbox{\cite{wolf-2020-transformers}} for BERT-base. Our \ccel model adopts two separate BiLSTMs, and the size of word-/character-level embedding is set to 100/30. The word-level embedding is initialized by training a fastText model on our CC corpus. 
All experiments are performed on a single NVIDIA 2080Ti graphics card (11GB).

\subsection{Evaluation Metrics}
We adopt the evaluation protocol of SemEval 2013 task 9.1 \cite{segura2013semeval} to evaluate models for mention extraction and entity linking. Precision, Recall, and F1 scores are reported in ``Partial'' mode (partial boundary match, regardless of the type) and ``Exact'' mode (exact boundary match, regardless of the type) for entity extraction. Besides, scores in ``Entity type'' mode (i.e., partial boundary match and correct entity type) are used for evaluating extraction and linking together.

\begin{table*}[!htbp]
\fontsize{10}{10}\selectfont
\centering
\begin{tabular}{lllccc}
\toprule
\textbf{\#} & \textbf{Extraction Model} & \textbf{Linking Model} & \textbf{Precision} & \textbf{Recall} & \textbf{F1} \\
\midrule
s.1 & \dag S\&M (S.1) & \dag S\&M (S.1) & \textbf{98.02} & 37.75 & 54.51 \\
s.2 & \dag S\&M (S.1 + S.2) & \dag S\&M (S.1 + S.2) & 79.34 & 45.02 & 57.44 \\
s.3 & \dag S\&M (S.1 + S.2 + S.3) & \dag S\&M (S.1 + S.2 + S.3) & 57.73 & 47.54 & 52.14 \\
\midrule
n.1 & \ccme-BERT (soft) & fastText (single-label) & 92.93 & 21.11 & 34.41 \\
n.2 & \ccme-BERT (soft) & fastText & 89.27 & 26.69 & 41.09 \\
n.3 & \ccme-BERT (soft) & BERT & 83.48 & 34.52 & 48.84 \\
n.4 & \ccme-BERT (soft) & \ccel & 84.36 & 45.22 & 58.88 \\
n.5 & \ccme-BERT (soft) & BERT-based Ranking & 82.74 & 46.51 & 59.55 \\
\midrule
b.1 & QuickUMLS & MedType (EHR) & 53.32 & 23.09 & 32.23 \\
m.1 & \dag S\&M (S.1 + S.2) & fastText (single-label) & 89.07 & 15.77 & 26.79 \\
m.2 & \dag S\&M (S.1 + S.2) & fastText & 86.49 & 19.52 & 31.85 \\
m.3 & \ccme-BERT (soft) & \dag S\&M (S.1) & \underline{97.86} & 45.60 & 62.20 \\
m.4 & \ccme-BERT (soft) & \dag S\&M (S.1 + S.2) & 85.64 & 51.53 & 64.34 \\
m.5 & \ccme-BERT (soft) & \dag S\&M (S.1) + \ccel & 86.28 & \underline{55.43} & \textbf{67.49} \\
m.6 & \ccme-BERT (soft) & \dag S\&M (S.1) + BERT-based Ranking & 85.48 & \textbf{55.61} & \underline{67.38} \\
\bottomrule
\end{tabular}
\vspace{-0.5em}
\caption{Entity linking performance. Scores are computed in \textbf{Entity Type} mode of SemEval'13. The \textbf{best}/\underline{2nd-best} scores in each column are in bold/underlined. \dag S.1, S.2, S.3 refer to string matching algorithms in Figure \ref{fig:framework}(a).}
\label{table:entity_type}
\vspace{-1em}
\end{table*}


\section{Results and Discussion}

\subsection{Results of Entity Mention Extraction}
\label{sec:result_ext}
Results of entity extraction are shown in table \ref{table:entity_extraction}. Among three matching-based methods, S\&M (HaPPy) has highest precision but lowest recall. This indicates that alternative strings provided by HaPPy ontology can lead to precise matches, but its coverage is very limited. Both S\&M (QuickUMLS) and S\&M (embedding) can achieve a better recall with the help of approximate string matching and embedding-based matching. But S\&M (embedding) also introduces a fairly high rate of false positives and leads to the worst precision.

Most neural models outperform the matching-based methods, indicating that machine learning models can learn task-relevant inductive bias from weak labels. \ccme-BERT performs better than LSTM, largely due to the better generalizability of BERT from pre-training. With the help of label smoothing, \ccme-BERT (soft) demonstrates better performance than \ccme-BERT model, suggesting that adjusting label weights by confidence can effectively alleviate the noise in weak labels. \ccme-ClinicalBERT, \ccme-BioBERT, \ccme-PubMedBERT, which were pre-trained with multiple biomedical related materials, perform slightly worse than the \ccme-BERT, suggesting the pre-training on biomedical corpus is not helpful for the entity extraction tasks on chief complaint corpus. Without loss of generality, we mainly use \ccme-BERT in following experiments.

We also consider the ensemble of \ccme-BERT models and matching methods. Given the output (extracted mention spans) of \ccme-BERT soft model, we further apply exact and approximate string matching aiming for a better precision.
Compared with \ccme-BERT (soft), both ensemble models achieve worse scores in terms of partial match, while \ccme-BERT (soft) with S\&M (HaPPy) performs the best for exact match. Although the overall performance is not improved, such ensemble models might useful when exact span matching is crucial to the task.



\subsection{Results of Entity Linking}
\label{sec:result_link}
We present the scores of entity linking in Table~\ref{table:entity_type}, 
Among three matching-based methods, S\&M (HaPPy) attains a high precision but a low recall. This confirms the observation from the extraction part. However, 
most neural models (n.1 to n.3) fail to beat the matching methods (s.1 to s.3) except for \ccel and BERT-based Ranking. Neural models can achieve a high precision score but the recall is relatively low. In the case of entity linking, weakly supervised labels cannot cover all the classes and most of them may concentrate on a small proportion of classes, thus models trained with weak labels may not perform well on classes that only have a small number of labels. Besides, different from entity extraction that models can utilize contextual information to detect spans, it is difficult for linking models to understand a target class if it is rarely seen in the data.


The major difference between \ccel and other neural models is the direct use of contextual information. 
The performance of \ccel drops from 58.88 to 51.04 (F1) after removing the context embedding part or the character embedding part, especially from 45.22 to 36.75 for recall. This confirms that including context information or character-level input can improve the model performance. Compared with \ccel, BERT-based Ranking model achieves higher recall but lower precision, indicating that the adding of entity titles as input can link better for unseen or rarely seen entities in the training set but also introduce more noises. As for fastText models, the single-label fastText model achieves highest precision among all neural models. Its advantage may lie in that the model is trained with the most likely labels and thus it is less affected by the noisy labels. Nonetheless, it fails to predict other concepts and results in the lowest recall.

Among all models, the general medical entity linking model MedType performs the worst, although it is pretrained with large medical datasets and annotated EHR documents. This indicates a non-negligible domain difference between chief complaint and common EHR data.
Among five combined models, using \ccme-BERT (soft) for extraction and S\&M for linking (m.3, m.4) outperforms the reverse ways (m.1, m.2) by a clear margin. This confirms that \ccme-BERT models are good at identifying entity mentions, while matching methods are good at classifying concepts given identified mentions. We also try experiments of replacing unreliable matching algorithms (S.2, S.3) with stacking \ccel or BERT-based Ranking on top of m.3 to handle the unmatched cases and it leads to a boost in recall.


\subsection{Effect of Weak Supervision}
\label{sec:result_weak_supervision}
In order to examine the effect of weak supervision, we conduct an ablation study by training models with and without weak labels.
To this end, we conduct a five-fold cross validation and report the average score of five runs, by taking 80\% of data points from the annotated test set as a training set for fully supervised learning, and the rest 20\% is used for testing. We also experiment with two training strategies: 1) Supervised: training models with annotated data only; 2) Fine-tuning: pre-train the model using weak labels and fine-tune it with the annotated data. The result in Table~\ref{table:entity_train_strategy} demonstrates that, even trained with little annotated data, \ccme-BERT can achieve decent results on mention extraction, 
but pre-training the model with weak labels can be beneficial for precision and F1.


\begin{table}[t]
\fontsize{9}{11}\selectfont
\centering
\begin{tabular}{lccc}
\toprule
\textbf{Training} & \textbf{P} & \textbf{R} & \textbf{F1} \\ 
\midrule
WeakSup & \textbf{83.41} & 56.70 & 67.51 \\
\midrule
Supervised & 77.76 & \textbf{89.66} & 83.29 \\
Fine-tuning & 82.25 & 85.98 & \textbf{84.07} \\
\bottomrule
\end{tabular}
\caption{Extraction performance (Partial Match) of \ccme-BERT (soft) with three training strategies.}
\label{table:entity_train_strategy}
\vspace{-1em}
\end{table}

\begin{table}[t]
\fontsize{10}{12}\selectfont
\centering
\begin{tabular}{lcc}
\toprule
\textbf{Train Data} & \textbf{Partial} & \textbf{Exact}\\
\midrule
w/ punct & 25.98 & 9.91 \\
w/o punct & 34.17 & 17.63 \\
w/ punct + denoising & \textbf{54.18} & \textbf{41.94} \\
\bottomrule
\end{tabular}
\caption{The extraction performance (F1) of \ccme-BERT (soft) model on the \textit{no-punctuation} test subset.}
\label{table:entity_partial_train_test}
\vspace{-1em}
\end{table}

\subsection{Resolving Cases without Punctuation Separators}
\label{sec:result_wo_punc}
We observe that in the chief complaint corpus nearly 40\% of chief complaint records do not contain any punctuation as separators. The split-and-match algorithm cannot work to provide weak labeled data, and the models may fail to generalize to this kind of cases. To prove this hypothesis, we split the training data to two parts by checking if it contains punctuation separators. We train \ccme-BERT with the two parts and evaluate models on a subset of the test set that only contains \textit{no-punctuation} instances.
The result in Table \ref{table:entity_partial_train_test} shows that the model trained with labels derived from \textit{with-punctuation} records does not generalize well to no-punctuation cases. 

To improve models' robustness against this issue, we propose to train the model with additional noise: we intentionally remove punctuation marks from the instances that contain them. In this way, the model is trained to ``denoise'' each record by implicitly predicting the existence of separators. The resulting model demonstrates significant improvements over the previous ones.


\subsection{Error Analysis}
The majority of errors can be grouped into two categories. First, the absence of separators in a record leads to many wrong predictions. For example, in the record of ``cough chest pain/congestion'', the separator to delimit ``cough'' and ``chest pain'' is missing. Thus S\&M methods are likely to miss one of the two chunks in mention extraction. Second, the variation in terminology can cause difficulty in linking. For example, we do not find the word ``dysuria'' in the dataset, since it's often referred to as ``urinary tract pain''. Even though neural models have the capability to infer synonyms, they can hardly perform well without a certain amount of annotated data to learn such knowledge.

\section{Conclusions}
We propose \ours, a weak supervision method for extracting and linking  chief complaint entities in the absence of human annotation. 
We develop a split-and-match algorithm to produce weak labels of both entity spans and concept labels. We also show that the customized label smoothing can effectively alleviate the noise in weak labels. 
Our framework is considered to be generic enough for chief complaint data across departments and we will check its generalizability on more data in the future. However, if domain of new dataset is different from ours, fine-tuning models on the new dataset may help to get better performance. If the format of new corpus is completely different from ours, e.g., they use different punctuation marks as separators, then the preprocessing steps and weak label generation steps need to change accordingly.
Our framework is also applicable to other entity identification tasks where human annotation is limited, and we leave it for future work. 
To further improve the performance of weak supervision, we will try heuristic methods \cite{vculjak2022strong} as well in the future.

\bibliography{anthology,custom}

\begin{thebibliography}{42}
\expandafter\ifx\csname natexlab\endcsname\relax\def\natexlab#1{#1}\fi

\bibitem[{Aronson(2001)}]{aronson2001effective}
Alan~R Aronson. 2001.
\newblock Effective mapping of biomedical text to the umls metathesaurus: the metamap program.
\newblock In \emph{Proceedings of the AMIA Symposium}, page~17. American Medical Informatics Association.

\bibitem[{Bhatia et~al.(2019)Bhatia, Celikkaya, Khalilia, and Senthivel}]{bhatia2019comprehend}
Parminder Bhatia, Busra Celikkaya, Mohammed Khalilia, and Selvan Senthivel. 2019.
\newblock Comprehend medical: a named entity recognition and relationship extraction web service.
\newblock In \emph{2019 18th IEEE International Conference On Machine Learning And Applications (ICMLA)}, pages 1844--1851. IEEE.

\bibitem[{Bing et~al.(2015)Bing, Chaudhari, Wang, and Cohen}]{bing2015improving}
Lidong Bing, Sneha Chaudhari, Richard~C Wang, and William Cohen. 2015.
\newblock Improving distant supervision for information extraction using label propagation through lists.
\newblock In \emph{Proceedings of the 2015 Conference on Empirical Methods in Natural Language Processing}, pages 524--529.

\bibitem[{Chang et~al.(2020)Chang, Hong, and Taylor}]{chang2020generating}
David Chang, Woo~Suk Hong, and Richard~Andrew Taylor. 2020.
\newblock Generating contextual embeddings for emergency department chief complaints.
\newblock \emph{JAMIA open}, 3(2):160--166.

\bibitem[{Chen et~al.(2021)Chen, Varoquaux, and Suchanek}]{Chen_Varoquaux_Suchanek_2021}
Lihu Chen, Gaël Varoquaux, and Fabian~M. Suchanek. 2021.
\newblock \href {https://ojs.aaai.org/index.php/AAAI/article/view/17499} {A lightweight neural model for biomedical entity linking}.
\newblock \emph{Proceedings of the AAAI Conference on Artificial Intelligence}, 35(14):12657--12665.

\bibitem[{Conway et~al.(2013)Conway, Dowling, and Chapman}]{conway2013using}
Mike Conway, John~N Dowling, and Wendy~W Chapman. 2013.
\newblock Using chief complaints for syndromic surveillance: a review of chief complaint based classifiers in north america.
\newblock \emph{Journal of biomedical informatics}, 46(4):734--743.

\bibitem[{{\v{C}}uljak et~al.(2022){\v{C}}uljak, Spitz, West, and Arora}]{vculjak2022strong}
Marko {\v{C}}uljak, Andreas Spitz, Robert West, and Akhil Arora. 2022.
\newblock Strong heuristics for named entity linking.
\newblock \emph{arXiv preprint arXiv:2207.02824}.

\bibitem[{Dara et~al.(2008)Dara, Dowling, Travers, Cooper, and Chapman}]{DARA2008613}
Jagan Dara, John~N. Dowling, Debbie Travers, Gregory~F. Cooper, and Wendy~W. Chapman. 2008.
\newblock \href {https://doi.org/https://doi.org/10.1016/j.jbi.2007.11.004} {Evaluation of preprocessing techniques for chief complaint classification}.
\newblock \emph{Journal of Biomedical Informatics}, 41(4):613--623.

\bibitem[{Dehghani et~al.(2017)Dehghani, Zamani, Severyn, Kamps, and Croft}]{dehghani2017neural}
Mostafa Dehghani, Hamed Zamani, Aliaksei Severyn, Jaap Kamps, and W~Bruce Croft. 2017.
\newblock Neural ranking models with weak supervision.
\newblock In \emph{Proceedings of the 40th International ACM SIGIR Conference on Research and Development in Information Retrieval}, pages 65--74.

\bibitem[{Duangsuwan and Saeku(2018)}]{duangsuwan2018semi}
Jarunee Duangsuwan and Pawin Saeku. 2018.
\newblock Semi-automatic classification based on icd code for thai text-based chief complaint by machine learning techniques.
\newblock \emph{International Journal of Future Computer and Communication}, 7(2).

\bibitem[{Gu et~al.(2021)Gu, Tinn, Cheng, Lucas, Usuyama, Liu, Naumann, Gao, and Poon}]{gu2021domain}
Yu~Gu, Robert Tinn, Hao Cheng, Michael Lucas, Naoto Usuyama, Xiaodong Liu, Tristan Naumann, Jianfeng Gao, and Hoifung Poon. 2021.
\newblock Domain-specific language model pretraining for biomedical natural language processing.
\newblock \emph{ACM Transactions on Computing for Healthcare (HEALTH)}, 3(1):1--23.

\bibitem[{Hofer et~al.(2018)Hofer, Kormilitzin, Goldberg, and Nevado-Holgado}]{hofer2018few}
Maximilian Hofer, Andrey Kormilitzin, Paul Goldberg, and Alejo Nevado-Holgado. 2018.
\newblock Few-shot learning for named entity recognition in medical text.
\newblock \emph{arXiv preprint arXiv:1811.05468}.

\bibitem[{Hoffmann et~al.(2011)Hoffmann, Zhang, Ling, Zettlemoyer, and Weld}]{hoffmann2011knowledge}
Raphael Hoffmann, Congle Zhang, Xiao Ling, Luke Zettlemoyer, and Daniel~S Weld. 2011.
\newblock Knowledge-based weak supervision for information extraction of overlapping relations.
\newblock In \emph{Proceedings of the 49th annual meeting of the association for computational linguistics: human language technologies}, pages 541--550.

\bibitem[{Horng et~al.(2019)Horng, Greenbaum, Nathanson, McClay, Goss, and Nielson}]{horng2019consensus}
Steven Horng, Nathaniel~R Greenbaum, Larry~A Nathanson, James~C McClay, Foster~R Goss, and Jeffrey~A Nielson. 2019.
\newblock Consensus development of a modern ontology of emergency department presenting problems—the hierarchical presenting problem ontology (happy).
\newblock \emph{Applied clinical informatics}, 10(03):409--420.

\bibitem[{Hsu et~al.(2020)Hsu, Weng, Wu, and Ho}]{hsu2020natural}
Jia-Hao Hsu, Ting-Chia Weng, Chung-Hsien Wu, and Tzong-Shiann Ho. 2020.
\newblock Natural language processing methods for detection of influenza-like illness from chief complaints.
\newblock In \emph{2020 Asia-Pacific Signal and Information Processing Association Annual Summit and Conference (APSIPA ASC)}, pages 1626--1630. IEEE.

\bibitem[{Hsu et~al.(2016)Hsu, Han, Arnold, Bui, and Enzmann}]{hsu2016data}
William Hsu, Simon~X Han, Corey~W Arnold, Alex~AT Bui, and Dieter~R Enzmann. 2016.
\newblock A data-driven approach for quality assessment of radiologic interpretations.
\newblock \emph{Journal of the American Medical Informatics Association}, 23(e1):e152--e156.

\bibitem[{Huang et~al.(2019)Huang, Altosaar, and Ranganath}]{huang2019clinicalbert}
Kexin Huang, Jaan Altosaar, and Rajesh Ranganath. 2019.
\newblock Clinicalbert: Modeling clinical notes and predicting hospital readmission.
\newblock \emph{arXiv preprint arXiv:1904.05342}.

\bibitem[{Ji et~al.(2020)Ji, Wei, and Xu}]{ji2020bert}
Zongcheng Ji, Qiang Wei, and Hua Xu. 2020.
\newblock Bert-based ranking for biomedical entity normalization.
\newblock \emph{AMIA Summits on Translational Science Proceedings}, 2020:269.

\bibitem[{Joulin et~al.(2016)Joulin, Grave, Bojanowski, Douze, J{\'e}gou, and Mikolov}]{joulin2016fasttext}
Armand Joulin, Edouard Grave, Piotr Bojanowski, Matthijs Douze, H{\'e}rve J{\'e}gou, and Tomas Mikolov. 2016.
\newblock Fasttext. zip: Compressing text classification models.
\newblock \emph{arXiv preprint arXiv:1612.03651}.

\bibitem[{Karagounis et~al.(2020)Karagounis, Sarkar, and Chen}]{Karagounis2020Coding}
Sotiris Karagounis, Indra~N. Sarkar, and Elizabeth~S. Chen. 2020.
\newblock Coding free-text chief complaints from a health information exchange: A preliminary study.
\newblock In \emph{AMIA Annual Symposium Proceedings}. American Medical Informatics Association.

\bibitem[{Kraljevic et~al.(2021)Kraljevic, Searle, Shek, Roguski, Noor, Bean, Mascio, Zhu, Folarin, Roberts, Bendayan, Richardson, Stewart, Shah, Wong, Ibrahim, Teo, and Dobson}]{Kraljevic2021-ln}
Zeljko Kraljevic, Thomas Searle, Anthony Shek, Lukasz Roguski, Kawsar Noor, Daniel Bean, Aurelie Mascio, Leilei Zhu, Amos~A Folarin, Angus Roberts, Rebecca Bendayan, Mark~P Richardson, Robert Stewart, Anoop~D Shah, Wai~Keong Wong, Zina Ibrahim, James~T Teo, and Richard J~B Dobson. 2021.
\newblock \href {https://doi.org/10.1016/j.artmed.2021.102083} {Multi-domain clinical natural language processing with {MedCAT}: The medical concept annotation toolkit}.
\newblock \emph{Artif. Intell. Med.}, 117:102083.

\bibitem[{Lee et~al.(2020)Lee, Yoon, Kim, Kim, Kim, So, and Kang}]{lee2020biobert}
Jinhyuk Lee, Wonjin Yoon, Sungdong Kim, Donghyeon Kim, Sunkyu Kim, Chan~Ho So, and Jaewoo Kang. 2020.
\newblock Biobert: a pre-trained biomedical language representation model for biomedical text mining.
\newblock \emph{Bioinformatics}, 36(4):1234--1240.

\bibitem[{Lee et~al.(2019)Lee, Levin, Finley, and Heilig}]{lee2019chief}
Scott~H Lee, Drew Levin, Patrick~D Finley, and Charles~M Heilig. 2019.
\newblock Chief complaint classification with recurrent neural networks.
\newblock \emph{Journal of biomedical informatics}, 93:103158.

\bibitem[{Li et~al.(2016)Li, Liu, Rastegar-Mojarad, Wang, Chaudhary, Therneau, and Liu}]{li2016topic}
Dingcheng Li, Sijia Liu, Majid Rastegar-Mojarad, Yanshan Wang, Vipin Chaudhary, Terry Therneau, and Hongfang Liu. 2016.
\newblock A topic-modeling based framework for drug-drug interaction classification from biomedical text.
\newblock In \emph{AMIA Annual Symposium Proceedings}, volume 2016, page 789. American Medical Informatics Association.

\bibitem[{Liao et~al.(2015)Liao, Cai, Savova, Murphy, Karlson, Ananthakrishnan, Gainer, Shaw, Xia, Szolovits et~al.}]{liao2015development}
Katherine~P Liao, Tianxi Cai, Guergana~K Savova, Shawn~N Murphy, Elizabeth~W Karlson, Ashwin~N Ananthakrishnan, Vivian~S Gainer, Stanley~Y Shaw, Zongqi Xia, Peter Szolovits, et~al. 2015.
\newblock Development of phenotype algorithms using electronic medical records and incorporating natural language processing.
\newblock \emph{bmj}, 350.

\bibitem[{N{\'e}v{\'e}ol et~al.(2017)N{\'e}v{\'e}ol, Zweigenbaum et~al.}]{neveol2017making}
Aur{\'e}lie N{\'e}v{\'e}ol, Pierre Zweigenbaum, et~al. 2017.
\newblock Making sense of big textual data for health care: findings from the section on clinical natural language processing.
\newblock \emph{Yearbook of medical informatics}, 26(1):228.

\bibitem[{Osborne et~al.(2020)Osborne, Booth, O'Leary, Mudano, Rosas, Foster, Saag, and Danila}]{Osborne2020Identification}
John~D. Osborne, James~S. Booth, Tobias O'Leary, Amy Mudano, Giovanna Rosas, Phillip Foster, Kenneth Saag, and Maria Danila. 2020.
\newblock Identification of gout flares in chief complaint text using natural language processing.
\newblock In \emph{AMIA Annual Symposium Proceedings}. American Medical Informatics Association.

\bibitem[{Rochefort et~al.(2015)Rochefort, Verma, Eguale, and Buckeridge}]{rochefort2015037}
C~Rochefort, A~Verma, T~Eguale, and D~Buckeridge. 2015.
\newblock O-037: surveillance of adverse events in elderly patients: a study on the accuracy of applying natural language processing techniques to electronic health record data.
\newblock \emph{European Geriatric Medicine}, (6):S15.

\bibitem[{Savova et~al.(2010)Savova, Masanz, Ogren, Zheng, Sohn, Kipper-Schuler, and Chute}]{savova2010mayo}
Guergana~K Savova, James~J Masanz, Philip~V Ogren, Jiaping Zheng, Sunghwan Sohn, Karin~C Kipper-Schuler, and Christopher~G Chute. 2010.
\newblock Mayo clinical text analysis and knowledge extraction system (ctakes): architecture, component evaluation and applications.
\newblock \emph{Journal of the American Medical Informatics Association}, 17(5):507--513.

\bibitem[{Segura~Bedmar et~al.(2013)Segura~Bedmar, Mart{\'\i}nez, and Herrero~Zazo}]{segura2013semeval}
Isabel Segura~Bedmar, Paloma Mart{\'\i}nez, and Mar{\'\i}a Herrero~Zazo. 2013.
\newblock Semeval-2013 task 9: Extraction of drug-drug interactions from biomedical texts (ddiextraction 2013).
\newblock Association for Computational Linguistics.

\bibitem[{Severyn and Moschitti(2015)}]{severyn2015twitter}
Aliaksei Severyn and Alessandro Moschitti. 2015.
\newblock Twitter sentiment analysis with deep convolutional neural networks.
\newblock In \emph{Proceedings of the 38th international ACM SIGIR conference on research and development in information retrieval}, pages 959--962.

\bibitem[{Shen et~al.(2017)Shen, Liu, Wang, Wang, Afzal, and Liu}]{shen2017leveraging}
Feichen Shen, Sijia Liu, Yanshan Wang, Liwei Wang, Naveed Afzal, and Hongfang Liu. 2017.
\newblock Leveraging collaborative filtering to accelerate rare disease diagnosis.
\newblock In \emph{AMIA Annual Symposium Proceedings}, volume 2017, page 1554. American Medical Informatics Association.

\bibitem[{Soldaini and Goharian(2016)}]{soldaini2016quickumls}
Luca Soldaini and Nazli Goharian. 2016.
\newblock Quickumls: a fast, unsupervised approach for medical concept extraction.
\newblock In \emph{MedIR workshop, sigir}, pages 1--4.

\bibitem[{Szegedy et~al.(2016)Szegedy, Vanhoucke, Ioffe, Shlens, and Wojna}]{szegedy2016rethinking}
Christian Szegedy, Vincent Vanhoucke, Sergey Ioffe, Jon Shlens, and Zbigniew Wojna. 2016.
\newblock Rethinking the inception architecture for computer vision.
\newblock In \emph{Proceedings of the IEEE conference on computer vision and pattern recognition}, pages 2818--2826.

\bibitem[{Teng et~al.(2020)Teng, Ma, Chen, Xiao, and Huang}]{teng2020automatic}
Fei Teng, Zheng Ma, Jie Chen, Ming Xiao, and Lufei Huang. 2020.
\newblock Automatic medical code assignment via deep learning approach for intelligent healthcare.
\newblock \emph{IEEE journal of biomedical and health informatics}, 24(9):2506--2515.

\bibitem[{Valmianski et~al.(2019)Valmianski, Goodwin, Finn, Khan, and Zisook}]{valmianski2019evaluating}
Ilya Valmianski, Caleb Goodwin, Ian~M Finn, Naqi Khan, and Daniel~S Zisook. 2019.
\newblock Evaluating robustness of language models for chief complaint extraction from patient-generated text.
\newblock \emph{arXiv preprint arXiv:1911.06915}.

\bibitem[{Vashishth et~al.(2020)Vashishth, Joshi, Dutt, Newman-Griffis, and Rose}]{vashishth2020medtype}
Shikhar Vashishth, Rishabh Joshi, Ritam Dutt, Denis Newman-Griffis, and Carolyn Rose. 2020.
\newblock Medtype: Improving medical entity linking with semantic type prediction.
\newblock \emph{arXiv preprint arXiv:2005.00460}.

\bibitem[{Wagner et~al.(2006)Wagner, Hogan, Chapman, and Gesteland}]{wagner2006chief}
Michael~M Wagner, William~R Hogan, Wendy~W Chapman, and Per~H Gesteland. 2006.
\newblock Chief complaints and icd codes.
\newblock \emph{Handbook of biosurveillance}, page 333.

\bibitem[{Wallace et~al.(2016)Wallace, Kuiper, Sharma, Zhu, and Marshall}]{wallace2016extracting}
Byron~C Wallace, Jo{\"e}l Kuiper, Aakash Sharma, Mingxi Zhu, and Iain~J Marshall. 2016.
\newblock Extracting pico sentences from clinical trial reports using supervised distant supervision.
\newblock \emph{The Journal of Machine Learning Research}, 17(1):4572--4596.

\bibitem[{Wolf et~al.(2020)Wolf, Debut, Sanh, Chaumond, Delangue, Moi, Cistac, Rault, Louf, Funtowicz, Davison, Shleifer, von Platen, Ma, Jernite, Plu, Xu, Scao, Gugger, Drame, Lhoest, and Rush}]{wolf-2020-transformers}
Thomas Wolf, Lysandre Debut, Victor Sanh, Julien Chaumond, Clement Delangue, Anthony Moi, Pierric Cistac, Tim Rault, Rémi Louf, Morgan Funtowicz, Joe Davison, Sam Shleifer, Patrick von Platen, Clara Ma, Yacine Jernite, Julien Plu, Canwen Xu, Teven~Le Scao, Sylvain Gugger, Mariama Drame, Quentin Lhoest, and Alexander~M. Rush. 2020.
\newblock \href {https://www.aclweb.org/anthology/2020.emnlp-demos.6} {Transformers: State-of-the-art natural language processing}.
\newblock In \emph{Proceedings of the 2020 Conference on Empirical Methods in Natural Language Processing: System Demonstrations}, pages 38--45, Online. Association for Computational Linguistics.

\bibitem[{Yang et~al.(2018)Yang, Liang, and Zhang}]{yang2018design}
Jie Yang, Shuailong Liang, and Yue Zhang. 2018.
\newblock Design challenges and misconceptions in neural sequence labeling.
\newblock In \emph{Proceedings of the 27th International Conference on Computational Linguistics}, pages 3879--3889.

\bibitem[{Zhao et~al.(2019)Zhao, Liu, Zhao, and Wang}]{zhao2019neural}
Sendong Zhao, Ting Liu, Sicheng Zhao, and Fei Wang. 2019.
\newblock A neural multi-task learning framework to jointly model medical named entity recognition and normalization.
\newblock In \emph{Proceedings of the AAAI Conference on Artificial Intelligence}, volume~33, pages 817--824.

\end{thebibliography}
\bibliographystyle{acl_natbib}

\end{document}